\documentclass{elsarticle}

\usepackage{amsmath,amssymb,amsfonts}
\usepackage{makecell}
\usepackage{algorithm}
\usepackage{algorithmic}
\usepackage{multirow}
\usepackage{diagbox}
\usepackage{color}
\journal{}

\begin{document}

\begin{frontmatter}



\title{ExperienceThinking: Constrained Hyperparameter Optimization based on Knowledge and Pruning}


\author[label1]{Chunnan Wang}
\ead{WangChunnan@hit.edu.cn}
\author[label1]{Hongzhi Wang\corref{cor1}}
\ead{hongzh@hit.edu.cn}
\author[label2]{Chang Zhou}
\ead{c7zhou@ucsd.edu}
\author[label1]{Hanxiao Chen}
\ead{1170400220@stu.hit.edu.cn}
\address[label1]{Harbin Institute of Technology, Harbin, China}
\address[label2]{University of California San Diego, California, USA}
\cortext[cor1]{Corresponding author}

\begin{abstract}
Machine learning algorithms are very sensitive to the hyperparameters, and their evaluations are generally expensive. Users desperately need intelligent methods to quickly optimize hyperparameter settings according to known evaluation information, and thus reduce computational cost and promote optimization efficiency. Motivated by this, we propose ExperienceThinking algorithm to quickly find the best possible hyperparameter configuration of machine learning algorithms within a few configuration evaluations. ExperienceThinking design two novel methods, which intelligently infer optimal configurations from two aspects: search space pruning and knowledge utilization respectively. Two methods complement each other and solve the constrained hyperparameter optimization problems effectively. To demonstrate the benefit of ExperienceThinking, we compare it with $3$ classical hyperparameter optimization algorithms with a small number of configuration evaluations. The experimental results present that our proposed algorithm provides superior results and achieve better performance.
\end{abstract}


\end{frontmatter}

\section{Introduction}\label{section:1}

Given a dataset $D$, an machine learning algorithm $A$ and $n$ hyperparameters $PN$=$\{P_1,$$\ldots$$,P_n\}$, the hyperparameter optimization (HPO) problem aims at finding an optimal configuration of $n$ hyperparameters, which maximizes the performance of $A$ in $D$. The HPO issues play an important role in the automatic machine learning (AutoML) field because most machine learning algorithms are black-box and they are very sensitive to the hyperparameter settings. A reasonable setting can promote the performance of machine learning model significantly. %HPO problems exist widely in the real life \color{red}{settings}\color{black}, and many common tasks in the computer science area, such as neural architecture search~\cite{b17,b18} and feature subset selection, can be transformed into and considered as such kind of problems.

To make machine learning algorithms deliver optimal performance in real applications, many HPO methods \cite{b9,b12,b13,b14,b15,c1} have been proposed. Among them, Grid Search, Random Search \cite{b2}, Genetic Algorithm \cite{b3,c3} and Bayesian Optimization \cite{b4,b5,c2} are very famous. Without taking various constraints into account, each of the existing HPO techniques can provide excellent solutions by traversing a large proportion of hyperparameter configurations. However, in practice, hyperparameter configuration space is generally complex and high-dimensional. Besides, in many cases, the evaluation of only one specific hyperparameter configuration can be extremely expensive for large models, complex machine learning pipelines, or large datasets \cite{b7,c4,c5}. Users are always unable to afford the huge expenses brought by the large numbers of configuration evaluations. However, well-performed hyperparameter configurations are still necessary to improve the performance of machine learning models. Therefore, they desperately need intelligent methods to help them find a good hyperparameter configuration with the limited budgets. Motivated by this, in this paper, we define a new problem, Constrained Hyperparameter Optimization (CHPO), as follows.

CHPO problem aims at finding a best possible hyperparameter configuration, which leads to great performance of the algorithm in the given dataset, utilizing a finite number of configuration evaluations. It allows users to put an upper limit on the number of configuration evaluations, according to their budget, which is more practical and user-friendly compared with HPO problem. However, this advantage also brings a crucial technical challenge that the configuration space is always very huge, whereas, the number of configuration evaluations are few and limited. It is not trivial to select the configurations to be estimated from such a huge space, and make sure that well-performed configurations are involved in such few candidates.

In this paper, facing this challenge, we carefully design \texttt{Human Experience} and \texttt{Parameter Analysis} approaches to analyze experience and intelligently infer optimal configurations, respectively, and thus increase the possibility of finding well-performed configurations.

Exploring inherent rules and finding knowledge from historical data can help us to better understand and solve problems. Based on this thought, we design \texttt{Human Experience}, a knowledge-driven approach, to find the optimal configurations with the help of knowledge. \texttt{Human Experience} extracts useful knowledge from known experience, and discover potential relation among configuration, configuration adjustment and the corresponding change of performance utilizing the obtained knowledge. It finally uses the discovered relation to infer optimal configurations reasonably. This method works well when most given hyperparameters are decisive for the performance. However, it may be less effective when most hyperparameters are redundant or unimportant to the performance, because much noise data may significantly influence the quality of the obtained knowledge and mislead it.

In order to solve this disadvantage, we develop \texttt{Parameter Analysis}, which applies pruning method, to cope with this challenge. \texttt{Parameter Analysis} analyzes the importance of each hyperparameter to the model performance, and reduces the configuration space by ignoring unimportant or redundant ones. Finally, it searches for the optimal configurations from much smaller space. Such method makes up for the limitation of \texttt{Human Experience}, because the space can be reduced significantly when most hyperparameters are redundant or unimportant, and this makes optimal configurations much easier to be found. Its shortcoming is that it may be less effective when most hyperparameters are decisive for the performance, because the adjusted space is very similar to the original one, and it is still very difficult to select optimal configurations from the new space. Obviously, such shortcoming could be overcome with \texttt{Human Experience}.

From above discussions, these two methods complement with each other. We combine them by developing respective advantage and finally propose a well-performed CHPO algorithm, which is called ExperienceThinking. In this paper, we also design a series of experiments to examine the ability of $3$ classic HPO algorithms to deal with CHPO problems, and compare with that of ExperienceThinking. The final results show that our proposed algorithm provides superior results and has better performance.

Major contributions of this paper are summarized as follows.

\begin{itemize}
\item Firstly, we propose CHPO problem, which is more practical and user-friendly than HPO problem.

\item Secondly, we develop two novel methods, i.e., \texttt{Human Experience} and \texttt{Parameter Analysis}, to intelligently infer optimal configurations from two different aspects.

\item Thirdly, we combine \texttt{Human Experience} and \texttt{Parameter Analysis}, and present the ExperienceThinking algorithm to effectively deal with CHPO problems.

\item Fourthly, We conduct extensive experiments to test the performance of ExperienceThinking and classic HPO algorithms for CHPO problems. The experimental results demonstrate the superiority of our proposed algorithm.
\end{itemize}

The remainder of this paper is organized into five sections. Section~\ref{section:2} introduces the existing HPO techniques. In Section~\ref{section:3}, we define the CHPO problem and some related concepts involved in this paper. Section~\ref{section:4} introduces \texttt{Human Experience} and \texttt{Parameter Analysis} approaches that we designed to analyze experience and intelligently infer optimal configurations. Section~\ref{section:5} gives our proposed algorithm ExperienceThinking. Section~\ref{section:6} compares and evaluates the ability of classic HPO techniques and ExperienceThinking to solve CHPO problem. Finally, we draw conclusions and present the future works in Section~\ref{section:7}.

\section{Related Work}\label{section:2}

Many modern methods and algorithms, e.g., deep learning methods and machine learning algorithms, are very sensitive to hyperparameters --- their performance functions are ``black-box'' and their performance depends more strongly than ever on the correct setting of many internal hyperparameters. In order to automatically find out suitable hyperparameter configurations, and thus promote the efficiency and effectiveness of the target method or algorithm, a number of HPO techniques have been proposed~\cite{b9,b10,b11,b12,b13,b14,b15}. In this section, we will provide a detailed introduction of three classic and commonly used HPO techniques, that are Grid Search, Random Search~\cite{b2} and Bayesian Optimization~\cite{b4,b5}, which are involved in our experimental part.

\textbf{Grid Search (GS).} GS is one of the most used and basic HPO methods in the literature. Each hyperparameter is discretized into a desired set of values to study, and GS evaluates the Cartesian product of these sets and finally chooses the best one as the optimal configuration. Although easy to implement, GS may suffer from the curse of dimensionality and thus become computationally infeasible, since the required number of configuration evaluations grows exponentially with the number of hyperparameters and the number of discrete levels of each. For example, 10 hyperparameters with 4 levels each would require 1,048,576 models to be trained. Even with a substantial cluster of compute resources, training so many models is prohibitive in most cases, especially with massive datasets and enormous calculations. 

\textbf{Random Search (RS).} RS is a simple yet surprisingly effective alternative of the GS. RS samples configurations at random until a certain budget for the search is exhausted, and chooses the best one as the optimal configuration. It explores the entire configuration space, and works better than GS when some hyperparameters are much more important than others~\cite{b2,b16}. However, its effectiveness is subject to the size and the uniformity of the sample. Candidate configurations can be concentrated in regions that completely omit the effective hyperparameter configurations, and it is likely to generate fewer improved configurations~\cite{b16}.

\textbf{Bayesian Optimization (BO).} BO is a state-of-the-art optimization method for the global optimization of expensive black box functions~\cite{b6}. BO works by fitting a probabilistic surrogate model to all observations of the target black box function made so far, and then using the predictive distribution of the probabilistic model, to decide which point to evaluate next. Finally, consider the tested point with the highest score as the solution for the given HPO problem. Different from GS and RS, which ignore historical observations, it makes full use of them to intelligently infer better configurations, and thus capable of providing better solutions within shorter time. Many works~\cite{b17,b18,a2,a3,a9} apply BO to optimize hyperparameters of neural networks due to its effectiveness. However, BO uses Gaussian Processes in background and like any other Gaussian Process model, it is going to scale poorly with number of hyperparameters, and will be very slow to converge when number of dimensions is big.

The ability of these three techniques to deal with HPO problems with a finite number of configuration evaluations has not been yet fully analyzed and systematically compared. In the experimental part, we make minor readjustments to these three techniques, making them suitable for dealing with various CHPO problems. We then analyze their performance with a certain finite number of estimates and compare with that of our proposed ExperienceThinking algorithm, in order to find out an effective method for dealing with CHPO problems. Details are shown in Section~\ref{section:6}.

\section{Problem Definition and Related Concepts}\label{section:3}

\subsection{CHPO Problem Definition}\label{section:3.1}

\textbf{Definition 1. (Constrained Hyperparameter Optimization Problem)} Consider a dataset $D$, a machine learning algorithm $A$, $n$ hyperparameters $PN$=$\{P_1,$$\ldots$$,P_n\}$, and an integer $N$. Let $\Lambda_{PN_i}$ denote the domain of $P_i$, $\Lambda_{PN}$=$\Lambda_{PN_1}$$\times$$\ldots$$\times$$\Lambda_{PN_n}$ denote the overall hyperparameter configuration space, and $f(\lambda,A,D)$ represent the performance score of machine learning algorithm $A$ in dataset $D$ under a hyperparameter configuration $\lambda \in \Lambda_{PN}$. The target of Constrained Hyperparameter Optimization (CHPO) problem is to find

\begin{equation}
\lambda^\ast = \mathop{argmax} \limits_{\lambda \in \Lambda_{PN}} {f(\lambda,A,D)}
\end{equation}

from $\Lambda_{PN}$, which maximizes the performance of $A$ in $D$, by evaluating $N$ configurations in $\Lambda_{PN}$. In this paper, we use the $P=(D,A,PN,N)$ quadruple to denote a CHPO problem.

\subsection{Related Concepts of CHPO}\label{section:3.2}

The following concepts are defined for the \texttt{Human Experience} approach description.

Consider a CHPO problem $P$=$(D,$$A,$$PN,$$N)$, where $D$ is a dataset, $A$ is a machine learning algorithm, $PN$ are $n$ hyperparameters and $N$ is an integer. We represent the overall hyperparameter configuration space as $\Lambda_{PN}$. A vector of hyperparameters is denoted by $\lambda$=$(\lambda_{(1)},$$\ldots$$,\lambda_{(n)})$$\in$$\Lambda_{PN}$, and the normalized version of $\lambda$ is denoted by $\overline{\lambda}\footnote{We use the following method to transform $\lambda$ into $\overline{\lambda}$: For numerical hyperparameters in $\lambda$, we apply Min-max normalization method to map its value to $[0,1]$; for textual ones, we replace its value with its index number in $\Lambda_{{PN}_i}$ first, and then apply the same method to map the value to $[0,1]$.}$. We use $f^\ast(A,D)$ to represent the ideal performance score of $A$ in $D$ ($f(\lambda,A,D)$$\le$$f^\ast(A,D)$, $\forall$$\lambda$$\in$$\Lambda_{PN}$). For example, $f^\ast(A,D)$ equals $1$ when $A$ is a classification model and classification accuracy is used to measure the model performance.

\textbf{Definition 2. (Configuration Difference, CDiffer)} Consider a CHPO problem $P$=$(D,$$A,$$PN,$$N)$, and two configurations $\lambda$,$\lambda^\prime$$\in$$\Lambda_{PN}$. The Configuration Difference (CDiffer) from $\lambda$ to $\lambda^\prime$ is defined as:
\begin{equation}
CDiffer(\lambda\rightarrow\lambda^\prime)=\lambda^\prime-\lambda
\end{equation}

\textbf{Definition 3. (Performance Difference, PDiffer)} Consider a CHPO problem $P$=$(D,$$A,$$PN,$$N)$, and two configurations $\lambda$,$\lambda^\prime$$\in$$\Lambda_{PN}$. The Performance Difference (PDiffer) from $\lambda$ to $\lambda^\prime$ is defined as:
\begin{equation}
PDiffer(\lambda\rightarrow\lambda^\prime)=\frac{f(\lambda^\prime,A,D)-f(\lambda,A,D)}{|f(\lambda,A,D)|}\times100\%
\end{equation}

\textbf{Definition 4. (Performance Promotion Space, PSpace)} Consider a CHPO problem $P$=$(D,$$A,$$PN,$$N)$, and a configuration $\lambda$$\in$$\Lambda_{PN}$. The Performance Promotion Space (PSpace) of $\lambda$ is defined as:
\begin{equation}
PSpace(\lambda)=\frac{f^\ast(A,D)-f(\lambda,A,D)}{|f(\lambda,A,D)|}\times100\%
\end{equation}

The smaller $PSpace(\lambda)$ is the better $\lambda$ is.

\textbf{Definition 5. (Ideal Adjustment, IAdjust)} Consider a CHPO problem $P$=$(D,$$A,$$PN,$$N)$, and a configuration $\lambda$$\in$$\Lambda_{PN}$. The Ideal Adjustment (IAdjust) of $\lambda$ is denoted as $IAdjust(\lambda)$, and the relationship between $IAdjust(\lambda)$ and $f^\ast(A,D)$ is as follows:
\begin{equation}
f(\lambda+IAdjust(\lambda),A,D)=f^\ast(A,D)
\end{equation}

and the relationship between $IAdjust(\lambda)$ and $PSpace(\lambda)$ is as follows:
\begin{equation}
PDiffer(\lambda\rightarrow\lambda+IAdjust(\lambda))=PSpace(\lambda)
\end{equation}

\section{Human Experience and Parameter Analysis}\label{section:4}

\texttt{Human Experience} and \texttt{Parameter Analysis} are the core of our proposed ExperienceThinking algorithm. Two  methods tell ExperienceThinking which configurations tend to be optimal by carefully analyzing and summarizing the historical experience, and thus guide ExperienceThinking to approach the global optimal configuration gradually. In this section, we will introduce these two intelligent methods in detail by revealing their internal operating mechanism.

\subsection{Human Experience Method}\label{section:4.1}

\textbf{Motivation.} Clearly, the knowledge of relations among configuration, configuration adjustment and the corresponding change of performance is helpful for solving the problem. Thus, we tend to design a knowledge-driven approach to find optimal configurations efficiently. Such approach brings two challenges. On the one hand, we need procedural knowledge to help us infer optimal configurations, while only factual knowledge (the performances of some configurations) is known. How to derive procedural knowledge from factual knowledge effectively is the problem to solve. On the other hand, the optimal configurations predicted by one model may not be completely trustworthy, due to the possible bias of single model, which is hard to avoid.

\begin{figure*}[t]
\centering
\includegraphics[width=\textwidth]{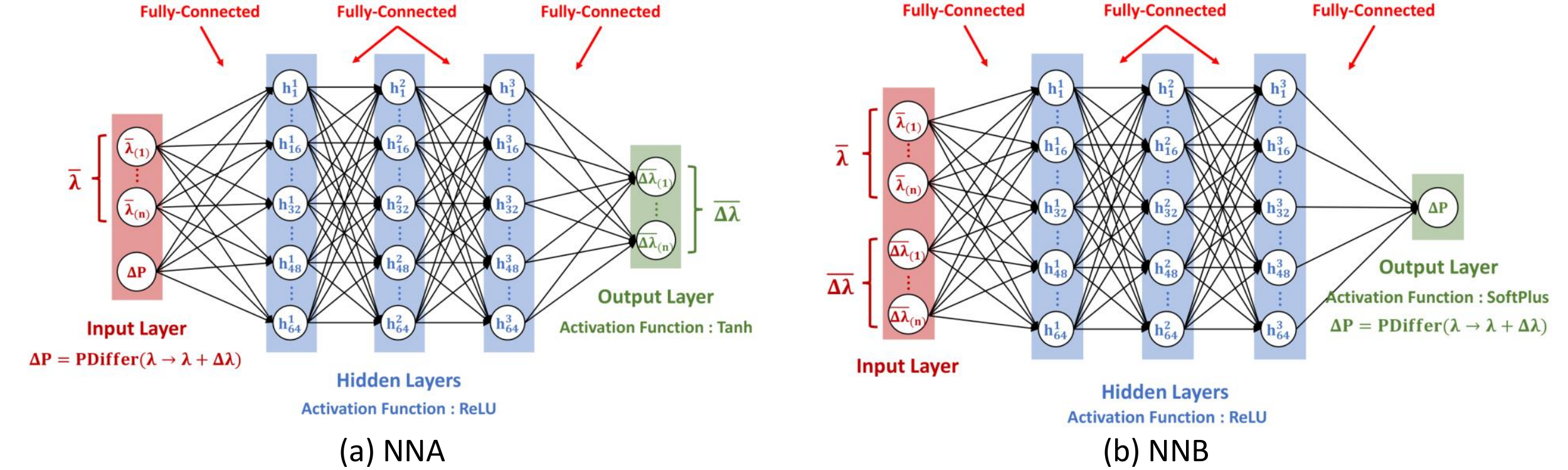}
\caption{The structures of two neural networks used in \texttt{Human Experience} method.}
\label{fig1}
\end{figure*}

\textbf{Design Idea.} Facing these two challenges, we develop knowledge representation and acquirement mechanism. We learn the procedural knowledge from the historical configurations with corresponding performance, which is a set of configuration-performance pairs, denoted as $Experience$. Consider two tuples $(\lambda,$$f(\lambda_i,$$A,$$D))$, $(\lambda^\prime,$$f(\lambda^\prime,$$A,$$D))$ $\in$ $Experience$. If $f(\lambda,$$A,$$D)$ $<$ $f(\lambda^\prime,$$A,$$D)$, then we consider that the performance of $\lambda$ can promote $\Delta P$=$PDiffer($$\lambda$$\rightarrow$$\lambda^\prime)$ if $\Delta\lambda$=$CDiffer($$\lambda$$\rightarrow$$\lambda^\prime)$ adjustment is made to $\lambda$ (i.e., $\lambda$ changes to $\lambda$+$\Delta\lambda$); instead, we say that the performance of $\lambda$ can decrease $\Delta P$ under $\Delta\lambda$ adjustment. Any two tuples in $Experience$ can provide us with two $($$\lambda$,$\Delta\lambda$,$\Delta P$$)$ triples as above, and we can obtain a total of $|Experience|$$\times$$|Experience$-$1|$ triples from $Experience$. These triples are useful for our understanding of the relationship among $\lambda$, $\Delta P$ and $\Delta\lambda$. We train neural networks with these triples, and consider the trained neural networks as the procedural knowledge, which assists us to find better configurations, e.g., setting $\Delta P$ to a high value.

To avoid the bias of single model, we also design multiple models to predict or verify optimal configurations, then ask them to discuss and exchange views, and thus improve the reliability of the predicted optimal configurations. We combine these solutions as \texttt{Human Experience} method as follows.

\begin{algorithm}[t]
\caption{HumanExperience}
\label{alg:HE}
\begin{algorithmic}[1]
\REQUIRE
a hyperparameter list $PN$=$\{P_1,$$\ldots$$,P_n\}$, an integer $Num$, a set of configuration-performance pairs $Experience$ = $\{(\lambda_i,$$f(\lambda_i,A,D))$ $|$ $i$=$1,$$\ldots$$t\}$
\ENSURE $Num$ optimal configuration candidates

\STATE NNAdjust $\gets$  build a neural network in Figure 1 (a)
\STATE $TrainData$ $\gets$ $\{$ $[$ $\overline{\lambda_i},$ $PDiffer(\lambda_j$$\rightarrow$$\lambda_i),$ \\$\overline{CDiffer(\lambda_j\rightarrow\lambda_i)}$ $]$ $|$ $i,j$ = $1,$$\ldots$$,t$ $\}$
\STATE $TrainData$ $\gets$ if the first two items of some tuples in $TrainData$ are the same, then only preserves one tuple among them, and remove the others from $TrainData$
\STATE train NNAdjust using $TrainDataA$ (inputs: the 1st and 2nd columns, outputs: the 3rd column, epochs: 300)
\STATE NNVerify $\gets$  build a neural network in Figure 1 (b)
\STATE train NNVerify using $TrainData$ (inputs: the 1st and 3rd columns, outputs: the 2nd column, epochs: 300)
\STATE $\overline{{{IAdjust}^\ast(\lambda_i)}}$ = NNAdjust$(\overline{\lambda_i},$$PSpace(\lambda_i))$, $i$=$1,$$\ldots$$,t$
\STATE ${PSpace}^\ast(\lambda_i)$ = NNVerify$(\overline{\lambda_i},$$\overline{{{IAdjust}^\ast(\lambda_i)}})$, $i$=$1,$$\ldots$$,t$
\STATE $Candidates$ $\gets$ $\{$$($ ${{IAdjust}^\ast(\lambda_i)}$+$\lambda_i$, $|PSpace(\lambda_i)$-${PSpace}^\ast(\lambda)|$ $)$ $|$ $i$ = $1,$$\ldots$$,t\}$
\STATE $Candidates$ $\gets$ delete tuples in $Candidates$ of which the 1st item $\in$ $\{\lambda_1,\ldots,\lambda_t\}$
\STATE sort the tuples in $Candidates$ in descending order of the 2nd item
\STATE $Optimals$ $\gets$ $\{$ $Candidates[i][0]$ $|$ $i$=$1,$$\ldots$$,Num$ $\}$
\RETURN $Optimals$
\end{algorithmic}
\end{algorithm}

\textbf{Detail Workflow.} Algorithm 1 shows the pseudo code of \texttt{Human Experience} method. Firstly, HumanExperience algorithm builds and trains NNAdjust, whose structure is shown in Figure 1(a), to fit the relationship between $(\overline{\lambda},$$\Delta P)$ and $\overline{\Delta\lambda}$ (Line 1-4). NNAdjust indicates what adjustment can be performed to make the performance of certain configuration achieve a certain increase, and thus help us infer optimal configurations. Note that in order to make gradient descent easier and convergence speed faster, $\lambda$ and $\Delta\lambda$ are normalized, $\overline{\lambda}$ and $\overline{\Delta\lambda}$ are used instead. Then, NNVerify, whose structure is shown in Figure 1(b), is built and trained to fit the relationship between $(\overline{\lambda},$$\overline{\Delta\lambda})$ and $\Delta P$ (Line 5-6). NNverify can tell how much the performance will increase if an adjustment is made to a certain configuration, and thus help us verify the effect of such certain adjustment. After obtaining these two well-trained neural networks, HumanExperience intelligently finds $Num$ optimal configuration candidates by utilizing them (Line 7-13). The details are as follows.

\textbf{Step 1.} Use NNAdjust to predict $\overline{IAdjust(\lambda)}$ aiming at inferring optimal configurations (Line 7). Taking $(\overline{\lambda},$$PSpace(\lambda))$ as input, NNAdjust outputs the predicted $\overline{IAdjust(\lambda)}$, which is denoted by $\overline{{IAdjust}^\ast(\lambda)}$. ${IAdjust}^\ast(\lambda)$+$\lambda$ is considered to be optimal according to NNAdjust.

\textbf{Step 2.} Use NNVerify to verify the rationality of $\overline{{IAdjust}^\ast(\lambda)}$ (Line 8). Taking $(\overline{\lambda},$$\overline{{IAdjust}^\ast(\lambda)})$ as input, NNVerify outputs the predicted $PDiffer(\lambda$$\rightarrow$ $\lambda$+${IAdjust}^\ast(\lambda))$, which is denoted by ${PSpace}^\ast(\lambda)$. ${PSpace}^\ast(\lambda)$ reflects NNVerify's view on the rationality of $\overline{{IAdjust}^\ast(\lambda)}$. More specifically, if ${PSpace}^\ast(\lambda)$ is very similar to $PSpace(\lambda)$, then we can say that both NNAdjust and NNVerify judge ${IAdjust}^\ast(\lambda)$+$\lambda$ to be optimal; otherwise, NNVerify disagrees with NNAdjust on the performance of ${IAdjust}^\ast(\lambda)$+$\lambda$.

\textbf{Step 3.} Select $Num$ configurations that are considered to be optimal by both of two neural networks (Line 9-13). The smaller $|PSpace(\lambda)$-${PSpace}^\ast(\lambda)|$ is, the more confidence NNAdjust and NNVerify have in the superiority of ${IAdjust}^\ast(\lambda)$+$\lambda$, and thus the more likely that ${IAdjust}^\ast(\lambda)$+$\lambda$ is optimal. Based on this methodology, $Candidates$ are sorted and $Num$ new configurations that are considered to be better are selected and output.

\textbf{Summary.} \texttt{Human Experience} extracts useful knowledge from the known configuration-performance information, and utilizes obtained knowledge to infer optimal configurations intelligently. Two neural networks used in it are like two human brains with different thought patterns \textemdash they infer or verify optimal configurations differently. They discuss with each other and exchange their views, and finally select the configuration candidates which are considered to be optimal by both of them. \texttt{Human Experience} brings forward a novel thought to infer optimal configurations.

As discussed in Section 1, \texttt{Human Experience} is a knowledge-driven method. Its effectiveness mainly depends on the quality of the obtained knowledge. It works well when most given hyperparameters are decisive for the performance. However, it may be less effective when most hyperparameters are redundant or unimportant to the algorithm performance, because much noise data are involoved and they may greatly influence the quality of the obtained knowledge and mislead it.

\subsection{Parameter Analysis Method}\label{section:4.2}

To make up for the limitation of \texttt{Human Experience}, we design \texttt{Parameter Analysis}, which brings forward another noval thought to find optimal configurations.

\textbf{Motivation.} Different hyperparameters may have different effects on algorithm performance. If we can figure out important hyperparameters utilizing $Experience$, then much better configurations are likely to be found. The reason is as follows. The opportunities to evaluate configurations are finite in CHPO problems, whereas, the configuration space is always huge. If we focus on important hyperparameters instead of unrelated or unimportant ones when deciding new configurations to test, then we can avoid wasting many evaluation opportunities on useless configurations, and have more opportunities to reach better and useful ones.

\textbf{Design Idea.} The key point in the implementation of the above idea is to judge the importance of hypeparameters reasonably. As we know, Random forest~\cite{b8} is a collection of decision trees, it can use the underlying trees in it to explain how each feature contributes to the model's predictive performance, and thus distinguishs the importance of features in the classification dataset. Thus, we can transfer the known experience into classification dataset first, then utilize the strong ability of random forest to judge the importance of hypeparameters for the search. Based on this idea, we design \texttt{Parameter Analysis} method.

\begin{algorithm}[t]
\caption{ParameterAnalysis}
\label{alg:PA}
\begin{algorithmic}[1]
\REQUIRE a hyperparameter list $PN$=$\{P_1,$$\ldots$$,P_n\}$, an integer $Num$, a set of configuration-performance pairs $Experience$ = $\{(\lambda_i,$$f(\lambda_i,A,D))$ $|$ $i$=$1$,$\ldots$,$t\}$
\ENSURE $Num$ optimal configuration candidates
\STATE sort the tuples in $Experience$ in ascending order of the second item
\STATE $psize=\lceil\frac{t}{3}\rceil$
\STATE $TrainData$ $\gets$ $\{$$(Experience[i][0],$$\lceil\frac{i}{psize}\rceil)$$|$$i$=$1,$$\ldots,$$t$$\}$
\STATE train the Random Forest classification model using $TrainData$, and obtain the importance score $Imp(P_i)$ of each hyperparameters $P_i$ $(i=1,$$\ldots,$$n)$
\STATE $Imps$ $\gets$ $\{$ $(P_i,$$Imp(P_i))$ $|$ $i$=$1,$$\ldots,$$n$ $\}$
\STATE sort tuples in $Imps$ in descending order of the 2nd item
\STATE $KeyPars$ $\gets$ $\emptyset$, $sum$ $\gets$ $0$, $i$ $\gets$ $0$
\WHILE{$sum$ $<$ $0.5$}
	\STATE $KeyPars$ $\gets$ $KeyPars$ $\bigcup$ $\{Imps[i][0]\}$
	\STATE $sum$ $\gets$ $sum$+$Imps[i][1]$, $i$ $\gets$ $i$+$1$
\ENDWHILE
\STATE $Optimals$ $\gets$ randomly generate $Num$ new configurations of $KeyPars$
\STATE $Optimals[i]$ $\gets$ the configuration of $KeyPars$ follows $Optimals[i]$ and that of other hyperparameters follows $Experience[t][0]$ $(i=1,$$\ldots,$$Num)$ \\ \# construct complete configurations of $PN$
\RETURN $Optimals$
\end{algorithmic}
\end{algorithm}

\textbf{Detail Workflow.} Algorithm 2 shows the pseudo code of \texttt{Parameter Analysis} method. Firstly, ParameterAnalysis algorithm converts $Experience$ into the classification dataset (Line 1-3). It ranks the configurations in the $Experience$ according to their performance, and classifies them into three categories, including high-performance ones (labeled by 3), mid-performance ones (labeled by 2) and low-performance ones (labeled by 1). In this way, each configuration has a category label related to their performance.

Secondly, ParameterAnalysis utilizes Random Forest to identify hyperparameters in $PN$ decisive for the performance (Line 4-11). In this step, ParameterAnalysis makes full use of the random forest to select key hyperparameters in $PN$, i.e., $KeyPars$, which have profound effects on the final performance.

Finally, ParameterAnalysis finds and outputs $Num$ optimal configuration candidates utilizing $KeyPars$ (Line 12-14). It generates $Num$ new configurations of $KeyPars$ randomly, and sets the values of other less important hyperparameters according to the most optimal configuration in $Experience$. In this way, $Num$ optimal configuration candidates of $PN$ are obtained.

\textbf{Summary.} \texttt{Parameter Analysis} applies pruning method. It utilizes Random Forest's strong ability of evaluating feature importance to reduce the configuration space, and thus improves the chances of finding optimal configurations.

This method complements with \texttt{Human Experience}. It works well especially when most given hyperparameters are redundant or unimportant, because the configuration space can be reduced a lot. However, it may be less effective when most hyperparameters are important, because the adjusted space is still very huge, and it is always very difficult to select the optimal configurations from the adjusted space.

\begin{figure}[t]
\centering
\includegraphics[width=0.8\columnwidth]{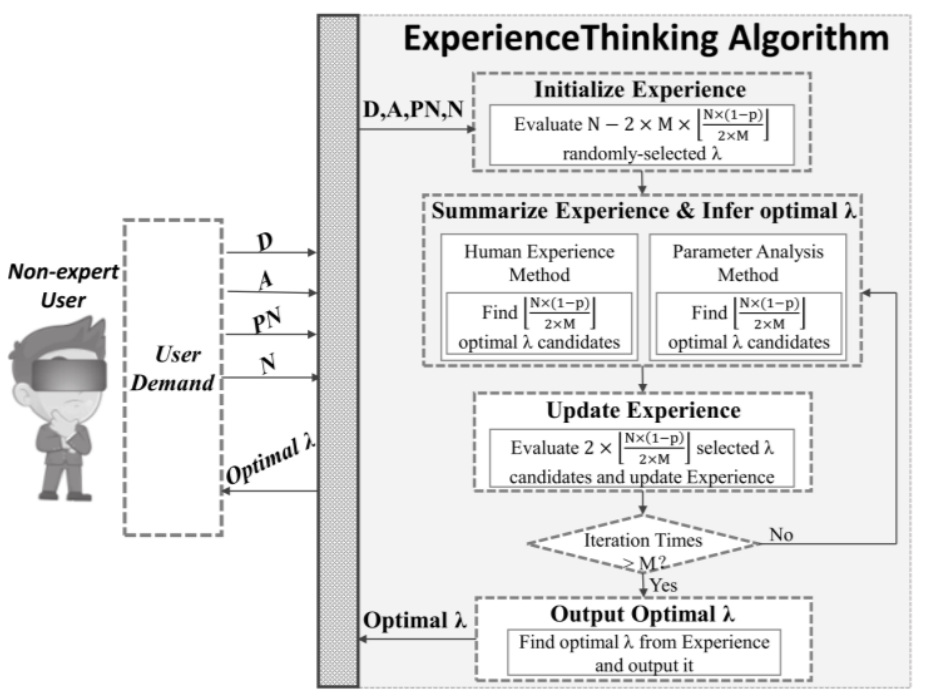}
\caption{The overall framework of ExperienceThinking. Given a CHPO problem $P$=$(D,$$A,$$PN,$$N)$, ExperienceThinking works as above and returns the optimal $\lambda$. Note that $M$ and $p$ are two parameters in ExperienceThinking, where 1$\leq$$M$$\leq$$\lfloor \frac{N\times(1-p)}{2}\rfloor$ represents the number of times 2 intelligent methods are invoked, and $0$$<$$p$$<$$1$ denotes the approximate percentage of configuration evaluations that are used to initialize experience, i.e. $\frac{N-2\times M\times \lfloor \frac{N\times(1-p)}{2\times M}\rfloor}{N}$ $\approx$ $p$.}
\label{fig2}
\end{figure}

\section{ExperienceThinking Algorithm}\label{section:5}

As discussed above, \texttt{Human Experience} and \texttt{Parameter Analysis} suit for different situations and complement each other. However, we are always unable to make the choice in advance. To increase the performance furthermore for various scenarios, we combine them and propose the ExperienceThinking algorithm. In the combined approach, these two approaches infer optimal configurations separately. All optimal candidate configurations provided by them are evaluated, and corresponding performance information is generated. $Experience$ is augmented with such configurations and performance. With the augmented $Experience$, these two approaches could constantly adjust themselves and enhance their credibility. Such adjustments are performed $M$ times, where $M$ is the constraint. The configuration with the best performance is considered as the solution for the given CHPO problem.
 Figure 2 is the overall framework of ExperienceThinking, and Algorithm 3 gives its pseudocode.

\textbf{Detail Workfolw.} For a CHPO problem, ExperienceThinking works as follows. Firstly, initializing experience, i.e., known configuration-performance pairs, by evaluating several randomly-selected $\lambda$ using about $p$ percent configuration evaluation opportunities (Line 1-2). Then, the iteration begins. ExperienceThinking divides the left configuration evaluation opportunities into $M$ parts equally. For each iteration, it invokes \texttt{Human Experience} method and \texttt{Parameter Analysis} method to analyze experience and infer optimal $\lambda$, after that utilizes a part of opportunities to evaluate several optimal $\lambda$ candidates provided by two intelligent methods and updates experience (Line 4-6). The iterative process is continued until the assumed number of evaluations is reached. Finally, the optimal $\lambda$ among $N$ evaluated configurations is output and considered as the solution for the given CHPO problem (Line 8-9).

%\footnote{how to determine $p$ and $M$ -- can't be deleted..., evaluated parameter sensitivity in experiment. other HPO techniques also have parameters, e.g. genatic algorithm and bayesian optimization.}
\begin{algorithm}[t]
\caption{ExperienceThinking}
\label{alg:ET}
\begin{algorithmic}[1]
\REQUIRE a CHPO problem $P$=$(D,$$A,$$PN,$$N)$, a percentage $p$, the limit $M$
\ENSURE a best possible configuration $\lambda$$\in$$\Lambda_{PN}$
\STATE $InitialConfs$ $\gets$ randomly select \\$N$-$2\times$$M\times$$\lfloor\frac{N\times (1-p)}{2\times M}\rfloor$ configurations from $\Lambda_{PN}$
\STATE $Experience$ $\gets$ $\{(\lambda,$$f(\lambda,A,D))$ $|$ $\lambda$ $\in$ $InitialConfs\}$
\FOR{$i$=1 to $M$}
	\STATE $NewConfs$ $\gets$ \\HumanExperience$(Experience,$$PN,$$\lfloor\frac{N\times (1-p)}{2\times M}\rfloor)$
	\STATE $NewConfs$ $\gets$ $NewConfs$ $\bigcup$ \\ParameterAnalysis$(Experience,$$PN,$$\lfloor\frac{N\times (1-p)}{2\times M}\rfloor)$
	\STATE $Experience$ $\gets$ $Experience$ $\bigcup$ \\$\{(\lambda,$$f(\lambda,A,D))$ $|$ $\lambda$ $\in$ $NewConfs\}$
\ENDFOR
\STATE $Optimal\lambda$ $\gets$ the most optimal $\lambda$ in $Experience$
\RETURN $Optimal\lambda$
\end{algorithmic}
\end{algorithm}

Note that these two methods may be run many times in ExperienceThinking. As the number of invocations grows, the experiences increase, these two methods become more reliable, and the configuration candidates suggested by them are more likely to be optimal. ExperienceThinking constantly adjusts two methods by enhancing their accuracy and thus gradually approaches the optimal configuration. This is just like the human growth processes\textemdash with the increase of their ages, humans accumulate richer experience and have stronger ability to solve problems, and the solution provided by them is improved. From this aspect, ExperienceThinking acts like a growing human and solve CHPO problems intelligently.

\section{Experiments}~\label{section:6}

To show the benefits of the proposed approach, we conduct extensive experiments using various CHPO problems. We implement all the algorithms in Python, and run experiments  irrelevant to CNN on an Intel 2.3GHz i5-7360U CPU machine with 16GB memory on Windows 10. As for the experiments related to CNN, we run them using GTX 1080 Ti.

Section~\ref{section:6.1} is the experimental setup, Section~\ref{section:6.2} gives the experimental results and Section~\ref{section:6.3} evaluates the parameter sensitivity and analyzes the importance of two methods in ExperienceThinking. 

\subsection{Experimental Setup}\label{section:6.1}

\textbf{Datasets.} We conduct experimental studies using 10 datasets, including 8 datasets used for data classification and 2 datasets used for image classification. Table~\ref{table1} shows the statistical information of them, and the following is the brief introduction to them.

\begin{table}[t]
\caption{10 datasets used in the experiments.}\smallskip
\footnotesize
\centering
\resizebox{0.95\columnwidth}{!}{
\smallskip\begin{tabular}{l l c c c}
\hline
Default Task & Datasets & No. of features & No. of classes & No. of records \\
\hline
\multirow{8}*{Data Classification} & zoo & 17 & 7 & 101 \\
& cbsonar & 60 & 2 & 208 \\
& image & 19 & 7 & 210 \\
& ecoli & 7 & 8 & 336 \\
& breast cancer & 30 & 2 & 569 \\
& balance & 4 & 3 & 625 \\
& creditapproval & 15 & 2 & 690 \\
& banknote & 4 & 2 & 1372 \\
\hline
\multirow{2}*{Image Classification} & cifar & 32x32x3 & 10 & 60,000 \\
& fashion mnist & 28x28x1 & 10 & 70,000 \\
\hline
\end{tabular}
}
\label{table1}
\end{table}

The first 8 datasets are available from UCI Machine Learning Repository\footnote{http://archive.ics.uci.edu/ml/datasets.php}. These datasets are from various areas, including life, computer, physics, society, finance and business. The cifar10 dataset\footnote{https://www.cs.toronto.edu/~kriz/cifar.html} is a collection of color images that are commonly used to train machine learning and computer vision algorithms --- consisting of a training set of 50,000 examples and a test set of 10,000 examples. It is collected by Alex Krizhevsky, Vinod Nair, and Geoffrey Hinton. The fashion mnist\footnote{https://research.zalando.com/welcome/mission/research-projects/fashion-mnist/} is a dataset of Zalando's article images-consisting of a training set of 60,000 examples and a test set of 10,000 examples. Each example is a 28x28 grayscale image.

\textbf{Algorithms for Comparison.} We implement three state-of-the-art HPO techniques: Grid Search (GS), Random Search (RS)~\cite{b2} and Bayesian Optimization (BO) with gaussian processes~\cite{a12}, which are introduced in Section~\ref{section:2}. We performed the following adjustments to these techniques making them suitable for dealing with CHPO problems and be able to compare with ExperienceThinking.

Consider a CHPO problem $P$=$(D,$$A,$$PN,$$N)$. RS randomly selects $N$ configurations from $\Lambda_{PN}$ and considers the optimal one among them as the solution to $P$. For each hyperparameter $P_i$, GS randomly select $\lfloor\sqrt[|PN|]{N}\rfloor$ or $\lceil\sqrt[|PN|]{N}\rceil$ values from $\Lambda_{{PN}_i}$, and thus form approximately $N$ (no more than $N$) configurations. GS evaluates these configurations and considers the most optimal one among them as the solution to $P$. BO randomly selects $\lfloor\frac{N}{2}\rfloor$ configurations from $\Lambda_{PN}$ as the initial samples, and then selects the next sample by optimizing acquisition function iteratively. The iteration stops when $N$-$\lfloor\frac{N}{2}\rfloor$ configuration evaluation opportunities are used up, and BO considers the most optimal evaluated configurations as the solution to $P$.

For ExperienceThinking algorithm, in the experiments, we set the parameter $p=0.5$ and $M=5$ by default. This setting will be demonstrated to be reasonable in the parameter sensitivity evaluation part.

\textbf{Evaluation Metrics.} In the experiments, if the hyperparameters $PN$ in the given CHPO problem $P$=$(D,$$A,$$PN,$$N)$ have the default configuration $\lambda_{def}$, then we use $PIRate$ to measure the effectiveness of the CHPO algorithm $S$, if $PN$ do not have $\lambda_{def}$, then we use $f(\lambda_{opt}^S,A,D)$ instead to quantify the effectiveness of $S$. For all efficiency experiments, we report the analysis time (all time cost except for time used for evaluating $N$ configurations) in minutes. The definition of $PIRate$ and the explanation of $f(\lambda_{opt}^S,A,D)$ are given as follows.

\textbf{Definition 6. (Performance Increase Rate, PIRate)} Consider a CHPO problem $P$=$(D,$$A,$$PN,$$N)$, and a CHPO technique $S$. Let $\lambda_{def}\in\Lambda_{PN}$ represent the default hyperparameter configuration, $\lambda_{opt}^{S}$$\in$$\Lambda_{PN}$ denote the optimal configuration of $P$ provided by $S$, and $f(\lambda,A,D)$ represent the performance score of machine learning algorithm $A$ in dataset $D$ under a hyperparameter configuration $\lambda \in \Lambda_{PN}$. The Performance Increase Rate (PIRate) of the algorithm in $P$ under $S$ is defined as:
\begin{equation}
PIRate(P,S)=\frac{f(\lambda_{opt}^S,A,D)-f(\lambda_{def},A,D)}{|f(\lambda_{def},A,D)|}\times100\%
\end{equation}

$PIRate(P,S)$ measures the performance difference between $\lambda_{opt}^S$ and $\lambda_{def}$. It can either be positive or negative. If $\lambda_{opt}^S$ outperforms $\lambda_{def}$, then $f(\lambda_{opt}^S,A,D)$ $>$ $f(\lambda_{def},A,D)$, and thus $PIRate(P,S)$ is positive; otherwise $PIRate(P,S)$ is negative. The higher $PIRate(P,S)$ value means the stronger ability of $S$ to solve $P$. Note that, in the following experiments, we divide $D$ into $3$ groups equally and apply 3-fold cross-validation accuracy to calculate $f(\lambda_{opt},A,D)$ and $f(\lambda_{def},A,D)$.

\subsection{Performance Evaluation}\label{section:6.2}

We examine the performance of ExperienceThinking, RS, GS and BO using three different types of CHPO problems, including CHPO problems related to the machine learning algorithm (Section~\ref{section:6.2.1}), CHPO problems related to neural architecture search (Section~\ref{section:6.2.2}, Section~\ref{section:6.2.3}) and CHPO problems related to feature subset selection (Section~\ref{section:6.2.4}). And we analyze all experimental results in Section~\ref{section:6.2.5}.

Note that for all CHPO problems, we run the CHPO algorithm 50 times by default, and report its average $PIRate$ (or average $f(\lambda_{opt}^S,A,D)$) and its average analysis time. Due to the fact that the analysis time of RS and GS is very little, we ignore them in the experiments.

\begin{table}[t]
\caption{Seven important hyperparameters in XGBoost.}\smallskip
\centering
\resizebox{.4\columnwidth}{!}{
\smallskip\begin{tabular}{l|r|r}
\hline
Name & Type & Set Ranges \\
\hline
n\_estimators & int & 10-200 \\
max\_depth & int & 5-20 \\
min\_child\_weight & int & 1-10 \\
gamma & float & 0.01-0.6 \\
subsample & float & 0.05-0.95 \\
colsample\_bytree & float & 0.05-0.95 \\
learning\_rate & float & 0.01-0.3 \\
\hline
\end{tabular}
}
\label{table2}
\end{table}

\subsubsection{XGBoost hyperparameter Optimization}\label{section:6.2.1}

\textbf{Experimental Design.} XGBoost (eXtreme Gradient Boosting)~\citep{b19} is a popular open-source implementation of the gradient boosted trees algorithm. From predicting ad click-through rates to classifying high energy physics events, XGBoost has proved its mettle in terms of performance and speed. It is very sensitive to hyperparameters --- its performance depends strongly on the correct setting of many internal hyperparameters. In this part, we try to automatically find out suitable hyperparameter configurations, and thus promote the effectiveness of XGBoost, utilizing CHPO techniques.

We consider seven main hyperparameters of XGBoost (shown in Table~\ref{table2}) as $PN$\footnote{https://xgboost.readthedocs.io/en/latest/ gives the default configuration of $PN$ ($\lambda_{def}$).}, set $N$ to 128 or 256, set $A$ to XGBoost algorithm and set $D$ to a data classification dataset in Table~\ref{table1}, and thus construct CHPO problems related to XGBoost to compare $4$ algorithms. Table~\ref{table3} show their performance.

\begin{table*}[t]
\caption{The average $PIRate$ and average analysis time of $4$ algorithms on CHPO problems related to XGBoost.}\smallskip
\centering
\resizebox{\textwidth}{!}{
\smallskip\begin{tabular}{l|r|r|r|r|r|r|r|r}
\hline
\multirow{2}{*}{Dataset} & \multicolumn{2}{c|}{RS} & \multicolumn{2}{c|}{GS} &
\multicolumn{2}{c|}{BO} & \multicolumn{2}{c}{ExperienceThinking}\\
\cline{2-9}
 & N=128 & N=256 & N=128 & N=256 & \multicolumn{1}{c|}{N=128} &\multicolumn{1}{c|}{N=256} & \multicolumn{1}{c|}{N=128} & \multicolumn{1}{c}{N=256} \\
\cline{1-9}
zoo & 0.66\% & 1.04\% & -5.45\% & -9.87\% & 1.07\%, \ 38.76 & 1.12\%, \ 124.78 & \textbf{1.23\%}, \ \textbf{6.83} & \textbf{1.73\%}, \ \textbf{24.02}\\
cbsonar & 30.40\% & 31.88\% & 22.67\% & 28.13\% & 29.25\%, \ 41.58 & 31.35\%, \ 158.61 & \textbf{32.17\%}, \ \textbf{7.08} & \textbf{37.26\%}, \ \textbf{23.48}\\
image & 5.45\% & 6.25\% & 0.36\% & 1.59\% & 5.27\%, \ 41.66 & 5.91\%, \ 263.52 & \textbf{7.77\%}, \ \textbf{7.40} & \textbf{9.43\%}, \ \textbf{25.70}\\
ecoli & 4.46\% & 4.69\% & 3.69\% & 3.88\% & 4.55\%, \ 31.40 & 4.77\%, \ 219.91 & \textbf{4.80\%}, \ \textbf{5.63} & \textbf{5.23\%}, \ \textbf{23.45}\\
breast cancer & 0.55\% & 0.71\% & -0.23\% & 0.06\% & 0.58\%, \ 36.09 & 0.62\%, \ 281.25 & \textbf{0.80\%}, \ \textbf{5.79} & \textbf{1.04\%}, \ \textbf{22.52}\\
balance & 7.03\% & 7.38\% & 6.36\% & 6.79\% & 7.58\%, \ 51.60 & 7.65\%, \ 143.47 & \textbf{7.68\%}, \ \textbf{6.41} & \textbf{7.85\%}, \ \textbf{23.92}\\
creditapproval & 2.43\% & 2.64\% & 1.55\% & 1.64\% & 2.57\%, \ 24.34 & 2.57\%, \ 201.34 & \textbf{2.68\%}, \ \textbf{6.05} & \textbf{2.75\%}, \ \textbf{23.93}\\
banknote & 0.14\% & 0.36\% & -0.40\% & -0.72\% & 0.46\%, \ 35.81 & 0.46\%, \ 210.51 & \textbf{0.56\%}, \ \textbf{7.05} & \textbf{0.60\%}, \ \textbf{24.43}\\
\cline{1-9}
Average Values & 6.39\% & 6.87\% & 3.57\% & 3.92\% & 6.42\%, \ 37.66 & 6.81\%, \ 200.40 & \textbf{7.21\%}, \ \textbf{6.53} & \textbf{8.24\%}, \ \textbf{23.93}\\
\hline
\end{tabular}
}
\label{table3}
\end{table*}

\textbf{Experimental Results.} From Table~\ref{table3}, we find that $4$ algorithms generally achieve higher $PIRate$ with the increase of $N$. ExperienceThinking is the most effective among them no matter what the value of $N$ is, GS performs the worst, the effectiveness of BO is slightly superior to that of RS. We also discover that BO and ExperienceThinking cost more time to analyze when $N$ gets larger, and the increase rate of analysis time cost by ExperienceThinking is much smaller than that of BO. ExperienceThinking outperforms BO no matter what the value of $N$ is.

\subsubsection{MLP Architecture Search}\label{section:6.2.2}

\textbf{Experimental Design.} Neural networks are powerful and flexible models that work well for many difficult learning tasks. Despite their success, they are still hard to design. In this part, we try to automatically design suitable MLP, a feedforward artificial neural network model, for the given dataset utilizing CHPO techniques.

\begin{table}[t]
\caption{Six important hyperparameters in MLP.}\smallskip
\centering
\resizebox{.7\columnwidth}{!}{
\smallskip\begin{tabular}{l|r|c}
\hline
Name & Type & Set Ranges or Available Options \\
\hline
hidden\_layer\_n & int & 1-20 \\
hidden\_layer\_size & int & 5-50 \\
activation & list & [relu, tanh, logistic, identity] \\
solver & list & [lbfgs, sgd, adam] \\
learning\_rate & list & [constant, invscaling, adaptive] \\
momentum & float & 0.1-0.9 \\
\hline
\end{tabular}
}
\label{table4}
\end{table}

We consider six main hyperparameters of MLP (shown in Table~\ref{table4}) as $PN$\footnote{https://scikit-learn.org/ gives the default configuration of $PN$ ($\lambda_{def}$).}, set $N$ to 128 or 256, set $A$ to MLP algorithm and set $D$ to one data classification dataset in Table~\ref{table1}, and thus construct several CHPO problems related to MLP architecture search to examine $4$ algorithms. Table~\ref{table5} shows their performance. We can see that ExperienceThinking recommends better hyperparameter configurations compared with other 3 methods, and its analysis time is far less than that of BO, in such CHPO problems.

\begin{table*}[t]
\caption{The average $PIRate$ and average analysis time of $4$ algorithms on CHPO problems related to MLP architecture search.}\smallskip
\centering
\resizebox{\textwidth}{!}{
\smallskip\begin{tabular}{l|r|r|r|r|r|r|r|r}
\hline
\multirow{2}{*}{Dataset} & \multicolumn{2}{c|}{RS} & \multicolumn{2}{c|}{GS} &
\multicolumn{2}{c|}{BO} & \multicolumn{2}{c}{ExperienceThinking}\\
\cline{2-9}
 & N=128 & N=256 & N=128 & N=256 & \multicolumn{1}{c|}{N=128} & \multicolumn{1}{c|}{N=256} & \multicolumn{1}{c|}{N=128} & \multicolumn{1}{c}{N=256} \\
\cline{1-9}
zoo & 1.41\% & 1.59\% & 0.91\% & 1.71\% & 1.46\%, \ 20.30 & 2.14\%, \ \ \ 63.55 & \textbf{1.62\%}, \ \textbf{7.60} & \textbf{2.49\%}, \ \textbf{23.43}\\
cbsonar & 129.22\% & 136.54\% & 87.37\% & 112.11\% & 105.75\%, \ 23.85 & 119.68\%, \ 133.40 & \textbf{136.18\%}, \ \textbf{6.93} & \textbf{138.64\%}, \ \textbf{25.07}\\
image & 116.29\% & 120.48\% & 101.90\% & 111.43\% & \textbf{120.57\%}, \ 43.85 & 120.95\%, \ 195.55 & \textbf{120.57\%}, \ \textbf{7.65} & \textbf{122.86\%}, \ \textbf{22.01}\\
ecoli & \textbf{33.20\%} & 35.00\% & 32.50\% & 33.19\% & 32.38\%, \ 28.11 & 35.05\%, \ 145.18 & \textbf{33.20\%}, \ \textbf{7.90} & \textbf{35.09\%}, \ \textbf{22.32}\\
breast cancer & 6.71\% & 7.06\% & 6.09\% & 6.65\% & \textbf{7.33\%}, \ 29.38 & 7.44\%, \ 130.94 & 7.28\%, \ \textbf{7.11} & \textbf{7.50\%}, \ \textbf{22.03}\\
balance & 10.96\% & 11.15\% & 9.57\% & 9.85\% & 10.98\%, \ 20.81 & 10.96\%, \ \ \ 87.95 & \textbf{11.40\%}, \ \textbf{7.47} & \textbf{11.73\%}, \ \textbf{23.92}\\
creditapproval & 13.36\% & 15.53\% & 7.20\% & 7.53\% & 16.15\%, \ 22.93 & 16.64\%, \ 161.05 & \textbf{16.75\%}, \ \textbf{7.69} & \textbf{17.10\%}, \ \textbf{23.15}\\
banknote & \textbf{0.00\%} & \textbf{0.00\%} & -0.18\% & -0.08\% & \textbf{0.00\%}, \ 14.94 & \textbf{0.00\%}, \ 120.64 & \textbf{0.00\%}, \ \textbf{6.36} & \textbf{0.00\%}, \ \textbf{21.76}\\
\cline{1-9}
Average Values & 38.89\% & 40.92\% & 30.67\% & 35.30\% & 36.83\%, \ 25.52 & 39.11\%, \ 129.78 & \textbf{40.88\%}, \ \textbf{7.34} & \textbf{41.93\%}, \ \textbf{22.96}\\
\hline
\end{tabular}
}
\label{table5}
\end{table*}

\begin{table}[t]
\caption{Fourteen important hyperparameters in CNN.}\smallskip
\scriptsize
\centering
\resizebox{\columnwidth}{!}{
\smallskip\begin{tabular}{l l p{4.5cm} p{4.5cm}}
\hline
Name & Type & Set Ranges or Available Options & Meaning\\
\hline
SL1Type & list & [Conv2D, MaxPooling2D, AveragePooling2D, Dropout] & The type of the 1st layer \\
SL2Type & list & [Conv2D, MaxPooling2D, AveragePooling2D, Dropout, None] & The type of the 2nd layer \\
SL3Type & list & [Conv2D, MaxPooling2D, AveragePooling2D, Dropout, None] & The type of the 3rd layer \\
SL4Type & list & [Conv2D, MaxPooling2D, AveragePooling2D, Dropout, None] & The type of the 4th layer \\
SL5Type & list & [Conv2D, MaxPooling2D, AveragePooling2D, Dropout, None] & The type of the 5th layer \\
SLActivation & list & [relu, softsign, softplus, selu, elu, softmax, tanh, sigmoid, hard$\_$sigmoid, linear] & The activation function used by the Conv2D layers in CNN \\
SLDroupout & float & 0.1$-$0.9 & The dropout rate set in Dropout layers used in the first five layers \\
DenseLNum & int & 0$-$3 & The number of fully connected layers in CNN \\
DenseLSize & list & [16,32,64,128,256,512,1024] & The number of neurons in each fully connected layer \\
DenseLDroupout & float & 0.1$-$0.9 & The dropout rate set in the Dropout layer used after fully connected layers \\
DenseLActivation & list & [relu, softsign, softplus, selu, elu, softmax, tanh, sigmoid, hard$\_$sigmoid, linear] & The activation function used by fully connected layers in CNN \\
OutputLActivation & list & [relu, softsign, softplus, selu, elu, softmax, tanh, sigmoid, hard$\_$sigmoid, linear] & The activation function used by the output layer in CNN \\
optimizer & list & [SGD, RMSprop, Adagrad, Adadelta, Adam, Adamax, Nadam] & The optimizer used by CNN \\
batch$\_$size & int & 10$-$100 & The batch size used when training CNN \\
\hline
\end{tabular}
}
\label{table6}
\end{table}

\subsubsection{CNN Architecture Search}\label{section:6.2.3}

\textbf{Experimental Design.} In this part, we try to automatically design suitable CNN, which is comprised of one or more convolutional layers (often with a subsampling step) and then followed by one or more fully connected layers as in a standard multilayer neural network, for the given image dataset utilizing CHPO techniques.

We consider fourteen hyperparameters related to CNN design (shown in Table~\ref{table6}) as $PN$, set $N$ to 128, set $A$ to CNN algorithm and set $D$ to one image classification dataset in Table~\ref{table1}, and thus construct several CHPO problems related to CNN architecture search to examine four CHPO algorithms. Table~\ref{table7} shows the performance of them.

\textbf{Experimental Results.} Since hyperparameters mentioned in Table~\ref{table6} do not have default values, we use $f(\lambda_{opt}^S,A,D)$ accuracy score to examine the effectiveness of the CHPO algorithm $S$. Besides, it is noticed that CNN training is very time-consuming, in order to save time, we set the epochs to 10 when training CNN, and run the CHPO algorithm 10 times to get average $f(\lambda_{opt}^S,A,D)$ and average analysis time. From Table~\ref{table7}, we find that ExperienceThinking performs the best among four algorithms, GS performs the worst, and RS performs better than BO. ExperienceThinking is more efficient than BO.

\begin{table*}[t]
\caption{The average $f(\lambda_{opt}^S,A,D)$ and average analysis time of $4$ algorithms on CHPO problems related to CNN architecture search.}\smallskip
\scriptsize
\centering
\resizebox{0.85\textwidth}{!}{
\smallskip\begin{tabular}{l|c c c c}
\hline
Dataset & RS & GS & BO & ExperienceThinking \\
\cline{1-5}
cifar10 & 0.660 & 0.383 & 0.613,\  88.37 & 0.673,\ 34.29 \\
fashion mnist & 0.878 & 0.627 & 0.864,\ 63.99 & 0.883,\ 34.43 \\
\cline{1-5}
Average Values & 0.769 & 0.505 & 0.739,\ 76.18 & 0.778,\ 34.36 \\
\hline
\end{tabular}
}
\label{table7}
\end{table*}

\subsubsection{Feature Subset Selection}\label{section:6.2.4}

\textbf{Experimental Design.} Feature subset selection is an important step in machine learning. Its idea is to find the best features that are suitable for the classification task. In this part, we use CHPO techniques to deal with feature subset selection problems.

We set $N$ to 128 or 256, set $A$ to K-Nearest Neighbor classification algorithm, set $D$ to one data classification dataset with more than 14 features in Table~\ref{table1}, and consider the features in $D$ as hyperparameters $PN$\footnote{Every three features construct a hyperparameter, where each feature corresponds to a value, that is 0 or 1. We preserve a feature in $D$ if the value of this feature is 1 in this experiment.} and thus construct several CHPO problems related to feature subset selection. We use four CHPO techniques to deal with these CHPO problems, and compare their performance. Note that, in this experiment, we consider the configuration of $PN$, which preserves all features in dataset $D$, as the default configuration $\lambda_{def}$. Table~\ref{table8} shows the results.

\textbf{Experimental Results.} The hyperparameters of CHPO issues in the previous three experiments are all fewer than 20. To further explore the performance of our algorithm ExperienceThinking, in this part we analyzed CHPO problems with more hyperparameters. For instance, the CHPO problem on cbsonar dataset that is analyzed in this part contains 60 hyperparameters. The experimental results in Table~\ref{table8} show that our method still outperforms other 3 methods on more hyperparameters. Note that BO cost too much time (more than 5 days) on the cbsonar dataset for getting average $PIRate$ and average analysis time, in this experiment. Since the time limit, we did not give the results of BO on cbsonar.

\begin{table*}[t]
\caption{The average $PIRate$ and average analysis time of $4$ algorithms on CHPO problems related to feature subset selection.}\smallskip
\centering
\resizebox{\textwidth}{!}{
\smallskip\begin{tabular}{l|r|r|r|r|r|r|r|r}
\hline
\multirow{2}{*}{Dataset} & \multicolumn{2}{c|}{RS} & \multicolumn{2}{c|}{GS} &
\multicolumn{2}{c|}{BO} & \multicolumn{2}{c}{ExperienceThinking}\\
\cline{2-9}
 & N=128 & N=256 & N=128 & N=256 & \multicolumn{1}{c|}{N=128} & \multicolumn{1}{c|}{N=256} & \multicolumn{1}{c|}{N=128} & \multicolumn{1}{c}{N=256} \\
\cline{1-9}
zoo & 	6.05\% & 6.77\% & 4.58\% & 4.93\% & 6.15\%, \ 36.38 & 6.48\%, \ 221.08 & \textbf{6.15\%}, \ \textbf{6.55} & \textbf{6.89\%}, \ \textbf{23.93}\\
cbsonar & 61.13\% & 64.03\% & 32.65\% & 35.33\% & $-$, \ $>$150 & $-$, \ $>$300\ \ & \textbf{78.46\%}, \ \textbf{7.35} & \textbf{85.88\%}, \ \textbf{23.74}\\
image & 41.91\% & 46.09\% & 23.48\% & 19.57\% & 43.48\%, \ 24.66 & 47.83\%, \ 235.88 & \textbf{45.91\%}, \ \textbf{6.69} & \textbf{50.65\%}, \ \textbf{21.19}\\
breast cancer & 2.13\% & 2.28\% & 	-0.34\% & -0.34\% & 2.13\%, \ 36.89 & 2.22\%, \ 226.38 & \textbf{2.16\%}, \ \textbf{6.35} & \textbf{2.29\%}, \ \textbf{23.67}\\
creditapproval & 39.10\% & 41.96\% & 27.34\% & 24.36\% & 39.80\%, \ 23.35 & 42.38\%, \ 154.05 & \textbf{41.53\%}, \ \textbf{7.82} & \textbf{43.78\%}, \ \textbf{27.92}\\
\cline{1-9}
Average Values & 30.06\% & 32.23\% & 17.54\% & 16.77\% & $-$, \ $-$\ \ \ \ \  & $-$, \ $-$\ \ \ \ \ \ \  & \textbf{34.84\%}, \ \textbf{6.95} & \textbf{37.90\%}, \ \textbf{24.09}\\
\hline
\end{tabular}
}
\label{table8}
\end{table*}

\subsubsection{Experimental Results Analysis}\label{section:6.2.5}

\textbf{Effectiveness Analysis.} The experimental results obtained from the above four experiments show us that the ability of ExperienceThinking to deal with CHPO problems is the strongest among four algorithms, GS performs the worst, and the effectiveness of BO is slightly superior than that of RS. Now let us analyze the reasons for different effectiveness performance of four algorithms.

For each hyperparameter, GS can only test very few values of it due to the limited number of configuration evaluations in CHPO problems. Besides, since these few values are randomly selected, not selected by domain experts, it is very likely that bad or ineffective configurations of the hyperparameter are selected, and thus result in the bad performance of GS. As for RS, although the tested values of each hyperparameter are also randomly selected, more values can be tested in RS. This makes RS more likely to find out better configurations and thus be more effectiveness compared with GS. However, GS and RS ignore historical observations and do not think deeply or analyze carefully for getting better configurations. This shortcoming makes GS and RS performs worse than BO and ExperienceThinking, which add intelligent analysis.

Both of BO and ExperienceThinking analyze historical experience intelligently for inferring better configurations, however, ExperienceThinking has two different analysis methods which complement each other, whereas BO only has one which uses Gaussian Processes in background. Due to the limited number of configuration evaluations in CHPO problems, the accuracy of each analysis module can not be guaranteed. Inferring optimal configurations with the help more reasonably designed analysis methods makes ExperienceThinking more reliable and thus be more effective compared with BO.

\textbf{Efficiency Analysis.} The experimental results obtained from the above four experiments show us that the analysis time of GS and RS is the smallest (can be ignored), and ExperienceThinking is far more efficient than BO. Now let us analyze the reasons for different efficiency performance of four algorithms.

GS and RS do not analyze historical experience and thus be more efficient than BO and ExperienceThinking, however, this also make them less effective. As for BO and ExperienceThinking, their analysis methods work differently and thus they have different time performance. The analysis methods in ExperienceThinking can provide many optimal configuration candidates at each iteration, and ExperienceThinking only need to invoke them several times (e.g., 5 set in the experiments) to get a good solution. However, the analysis module used in BO can only provide one candidate each time, and BO need to invoke it many times. Two analysis methods used in ExperienceThinking are not time-consuming, besides, they are invoked very few times, therefore the time performance of ExperienceThinking is far more efficient than BO.

\textbf{Summary.} Since the configuration evaluations in CHPO problems are commonly very expensive and time-consuming, users do not want to get a inferior solution after evaluating configurations. If much better solutions can be obtained at the cost of a certain amount of time for analyzing, users would gladly agree. Though ExperienceThinking is less efficient than GS and RS, but its effectiveness is the highest, besides, it efficiency is acceptable (better than BO), therefore, ExperienceThinking is the best CHPO algorithm among four algorithms that we analyzed.

\subsection{Parameter Sensitivity and Module Importance Evaluation}\label{section:6.3}

In this part, we examine the effect of two parameters on the performance of ExperienceThinking (Section~\ref{section:6.3.1}), and analyze the importance of two methods in ExperienceThinking (Section~\ref{section:6.3.2}).

\subsubsection{Parameter Sensitivity Evaluation}\label{section:6.3.1}

We investigate the effect of $p$ and $M$ on the performance of ExperienceThinking using CHPO problems analyzed above. Table~\ref{table9} is an example on a CHPO problem $P$=(cbsonar,XGBoost,$PN$,128), where $PN$ consists of seven hyperparameters in Table~\ref{table2}. 

As we can see, the $PIRate$ increases first and then decreases with the increasing of $p$, and the analysis time increases with the increase of $p$. The reasons are as follows. When $p$ is very big, ExperienceThinking is very similar to RS, which ignores the historical information, and thus be ineffective. When $p$ is very small, the initial few configurations can be concentrated in regions that completely omit the effective hyperparameter configuration and thus be useless for inferring better configurations, thus forming a vicious circle. This makes ExperienceThinking ineffective. Besides, with the increase of $p$, more experience is considered in the two analysis methods, and thus makes the analysis time used by ExperienceThinking longer. For getting a better solution, we suggest users to set $p$ to 0.5 when using ExperienceThinking.

\begin{table}[t]
\caption{Varying $p$ or $M$: average $PIRate$ and average analysis time.}\smallskip
\centering
\resizebox{0.85\columnwidth}{!}{
\smallskip\begin{tabular}{l|r|r|r|r|r|r}
\hline
\multirow{2}{*}{Performance} & \multicolumn{3}{c|}{Varying $p$ ($M$=3)} & \multicolumn{3}{c}{Varying $M$ ($p$=0.3)}\\
\cline{2-7}
 & $p$=0.1 & $p$=0.5 & $p$=0.9 & $M$=1 & $M$=5 & $M$=10 \\
\cline{1-7}
$PIRate$ & 32.16\% & 33.29\% & 31.01\% & 31.47\% & 33.12\% & 33.82\%\\
analysis time & 1.76 & 3.94 & 6.69 & 0.32 & 5.67 & 13.51 \\
\hline
\end{tabular}
}
\label{table9}
\end{table}

As for $M$, the $PIRate$ and the analysis time increase with the increasing of $M$. The reasons are as follows. More adjustments are made to improve the reliability of two analysis methods in ExperienceThinking, with the increase of $M$. This makes the configuration candidates suggested by two methods are more likely to be optimal and thus enhance the effectiveness of ExperienceThinking. However, more invocations mean much more analysis time, and this makes ExperienceThinking less efficient. For getting a better solution, we suggest users to set $M$ as large as possible when using ExperienceThinking.

\subsubsection{Module Importance Evaluation}\label{section:6.3.2}

Since the design of two mudules \texttt{Human Expereince} and \texttt{Parameter Analysis} complement with each other well in ExperienceThinking, it seems necessary for us to do some experiments and verify the functions of these to procedures.
Thus,we remove Human Experience part or Parameter Analysis part from our algorithm, and thus construct two algorithms HEA and PAA. We use them to solve the CHPO problems related to XGBoost, where the budget N is set to 128 and dataset D is set to image or ecoli, and thus examine their functions. The final results show that the PIRate of two algorithms (HEA: 5.76\% on image, 4.10\% on ecoli; PAA: 5.68\% on image, 4.62\% on ecoli) are lower than that of their combination (7.77\% on image, 4.80\% on ecoli). This proves that two procedures are complementary (they work as our paper tells indeed), and combining them in our method is reasonable.

\section{Conclusion and Future Works}\label{section:7}

In this paper, we present and formulate the CHPO problem, which aims at dealing with HPO problem as effectively as possible under limited computing resource. Compared with classic HPO problem, CHPO problem is more practical and user-friendly. Besides, we fully utilize the knowledge extracted from the historical experience and combine the pruning strategy, and thus propose an effective algorithm ExperienceThinking to solve CHPO problem. We also design a series of experiments to examine the ability of three classic HPO techniques to deal with CHPO problems, and compare with that of ExperienceThinking. The extensive experimental results show that our proposed algorithm provides more superior results and has better performance on various CHPO problems. In the future works, we will try to design effective methods to reduce the computational cost of model evaluation, and thus further improve the efficiency of our approach.

%\section*{Acknowledgements}

%This paper was partially supported by NSFC grant U1509216, U1866602, 61602129, 61772157,  CCF-Huawei Database System Innovation Research Plan DBIR2019005B and Microsoft Research Asia.



%%       \citet{<label>} ==> Jones et al. [21]
%%       \citep{<label>} ==> [21]
%%



\bibliographystyle{elsarticle-num}
\bibliography{ref}

%%
%% End of file `elsarticle-template-num-names.tex'.

\end{document}